\documentclass{article}

\usepackage{PRIMEarxiv}

\usepackage[utf8]{inputenc} 
\usepackage[T1]{fontenc}    
\usepackage[colorlinks=true, linkcolor=black, citecolor=black, urlcolor=black]{hyperref}
\usepackage{url}            
\usepackage{booktabs}       
\usepackage{amsfonts}       
\usepackage{nicefrac}       
\usepackage{microtype}      
\usepackage{lipsum}
\usepackage{fancyhdr}       
\usepackage{graphicx}       
\graphicspath{{media/}}     

\title{

Quantitative Evaluation of KIRETT Wearable Demonstrator for Rescue Operations

\thanks{\textit{2024 IEEE World AI IoT Congress (AIIoT) | 979-8-3503-8780-3/24/\$31.00 ©2024 IEEE | DOI: 10.1109/AIIoT61789.2024.10578968
}}}

\author{
  Mubaris Nadeem, Johannes Zenkert, Lisa Bender, Christian Weber, Madjid Fathi \\
  Institute for Knowledge-Based Systems and Knowledge Management \\
  University of Siegen\\
  Hoelderlinstrasse 3, 57068 Siegen, Germany\\
  \texttt{\{Mubaris Nadeem\}mubaris.nadeem@uni-siegen.de} \\
}

\begin{document}
\maketitle

\begin{abstract}

Healthcare and Medicine are under constant pressure to provide patient-driven medical expertise to ensure a fast and accurate treatment of the patient. In such scenarios, the diagnosis contains, the family history, long term medical data and a detailed consultation with the patient. In time-critical emergencies, such conversation and time-consuming elaboration are not possible. Rescue services need to provide fast, reliable treatments for the patient in need. With the help of modern technologies, like treatment recommendations, realtime vitals-monitoring, and situation detection through artificial intelligence (AI) a situation can be analyzed and supported in providing fast, accurate patient-data-driven medical treatments. In KIRETT, a wearable device is developed to support in such scenarios and presents a way to provide treatment recommendation in rescue services. The objective of this paper is to present the quantitative results of a two-day KIRETT evaluation (14 participants) to analyze the needs of rescue operators in healthcare.

\end{abstract}


\section{Introduction}
The emergency services are under constant pressure to make medical diagnoses quickly with little to prior knowledge of the situation at hand and to support and stabilize a patients' health with the right treatment \cite{zenkert_kirett_2022}. Characteristics such as age, weight and vital signs are essential to apply the appropriate treatment and track the patient's state of health, till its successful transfer to the hospital. Due to a variety of circumstances, such as a lack of time and information, it is possible to analyze for diagnoses which does not apply. This can lead to incorrect decisions in such potentially critical situations. With the help of modern digitalization measures and the use of artificial intelligence, rescue operations can be technologically supported by generating recommendations for treatment \cite{lwda_kirett} and to provide illness groups, based on historical data \cite{cardioShad}. To investigate this, the KIRETT project was designed to evaluate whether the use of treatment recommendations and AI in rescue operations in the form of a wearable device can support rescue operations in their daily work. Such wearable devices, provides a local IoT environment, in which multiple medical devices are connected. This paper describes the quantitative results of the two-day user study within the KIRETT project.

\begin{figure*}[htbp]
\centerline{\includegraphics[width=\textwidth]{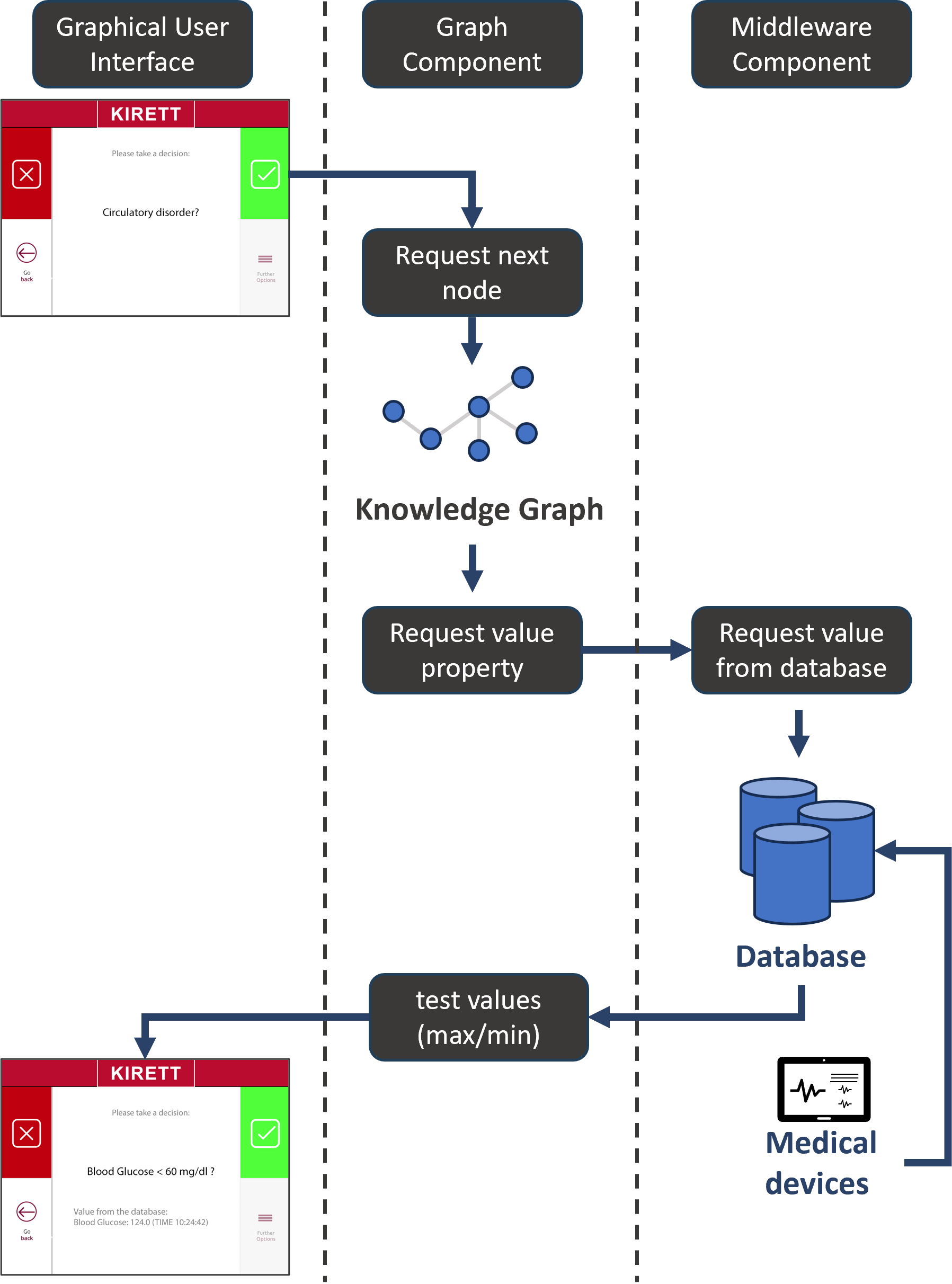}}
\caption{\textbf{Setup of the KDT:} A test group of two rescuers (left), will initially be provided an introduction and conduct a pre-interview with regards on personal information and their experience with artificial intelligence. Afterwards they have n-many iterations of various test-cases exploring the KIRETT-demonstrator. Subsequently, a qualitative interview follows, discussion various experiences gained in the testing phase. A Hardware-Table is provided to check and elaborate the KDT-hardware with a questionnaire at the end.}
\label{generalsetup}
\end{figure*}

\section{Background}
\subsection{The KIRETT Project}
The KIRETT project describes a local Internet of Things (IoT) application in which several interfaces to physical devices were combined with software-based components. Embedded systems were used, which were connected to medically certified end devices with the help of Bluetooth communication interfaces in order to transmit vital parameters and important medical core variables. The aim was to provide recommendations for treatments and to recognize the situation \cite{cardioShad} in order to evaluate the current circumstances in real-time. The medical decision-support using the recommended action is based on a knowledge graph \cite{lwda_kirett} constructed for the project. Communication between the various modules was made possible with a communication system. A wearable device was designed to investigate, how the UX/UI for rescue operations needs to be designed \cite{uxNadeem}. 

The project uses a Field Programmable Gate Array (FPGA) as a key component for hardware-assisted situational awareness \cite{ezekiel2023time}. In the field of healthcare and emergency medicine, it is essential to use innovative technologies to make rescue services more efficient and effective. The FPGA provides the ability to execute complex real-time situational awareness algorithms directly at the hardware level, resulting in a significant acceleration of analysis processes \cite{ezekiel2023optimization}. By utilizing these hardware-accelerated approaches, critical information can be extracted and transmitted to emergency responders faster, which in turn improves response time and treatment accuracy.

\subsection{Related Work}
With regard to the evaluation of the KIRETT wearable, references can be made to other studies that deal with a similar topic. Yin et al. \cite{yin2022user} investigated the acceptance of smart medical wearables and the factors that influence usage behavior using the unified theory of acceptance and use of technology (UTAUT) and the technology acceptance model (TAM). This study aims to expand the current understanding of smart medical wearables and examine the factors that influence their adoption. The results can help contextualize the findings of the KIRETT evaluation by providing insights into end-user needs and preferences.

Brander and von Schewen Sterndal \cite{brander2014development} describe the development of a wearable health device specifically designed to meet the needs of nurses. Through intensive user involvement in the development process, a wearable alarm system called ELSA was designed to improve the efficiency and safety of hospital work. This work provides valuable insights into the process of user-centrality in the development process, which was also focused on in the KIRETT project.

\section{Methodology}
\subsection{Study design}
\begin{figure}[htbp]
\centerline{\includegraphics[width=250px]{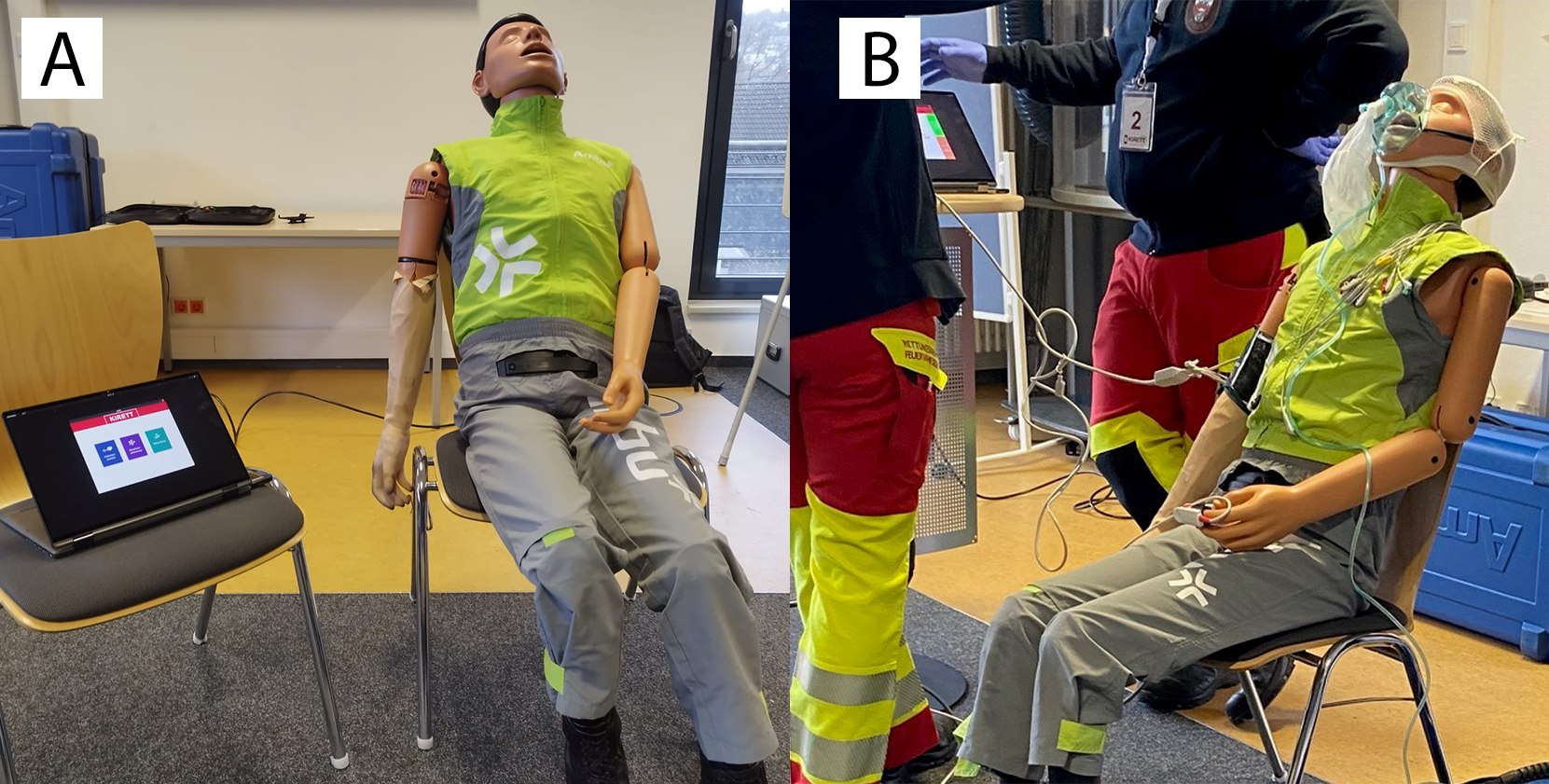}}
\caption{\textbf{Setup of the KDT:} The dummy with the sofware-based KIRETT-demonstrator. A shows the demonstrator setup and B the active usecase-treatment with rescue operators.}
\label{dummy}
\end{figure}

For the KIRETT-Demonstrator-Testing (KDT) a two-day evaluation at the Fire station in Siegen, Germany was planned. After a 90 minutes hands-on evaluation, a qualitative expert-interview and a quantitative survey was conducted. The testing was run between the 29th of February 2024 and the 1st of March 2024. No recall was planned for the evaluation. For the KDT a combined evaluation: qualitative evaluation (Fig. \ref{generalsetup}, Experts-interview) and quantitative evaluation (Fig. \ref{generalsetup}, Questionnaire) was constructed to gain the maximum possible insights for the central research questions of the project. Figure \ref{generalsetup} describes the study design for the evaluation conducted. Two participants are randomly matched together and guided through the test-setting. Starting with a brief introduction about the project and the test-setup, a first questionnaire was conducted. This form holds six questions with single and multiple choice, free-text and Likert-scaled answer options about the respondent's own person and personal experience with artificial intelligence, as well as their opinion on the integration of artificial intelligence in the emergency services. Subsequently the participants had the opportunity to work actively for 60 to 90 minutes with the testing device (Fig. \ref{dummy}, A). Various emergency use-cases were coordinated through a moderator. For this purpose, a software-based solution of the demonstrator was integrated into a learning environment familiar to the testers to reduce new impressions that were irrelevant for the testing.  Six test rescue missions were constructed beforehand, of which 2-3 test-cases were sent to the test groups (Fig. \ref{dummy}, B).

\begin{figure*} [htbp]
    \centerline{\includegraphics[width=\linewidth]{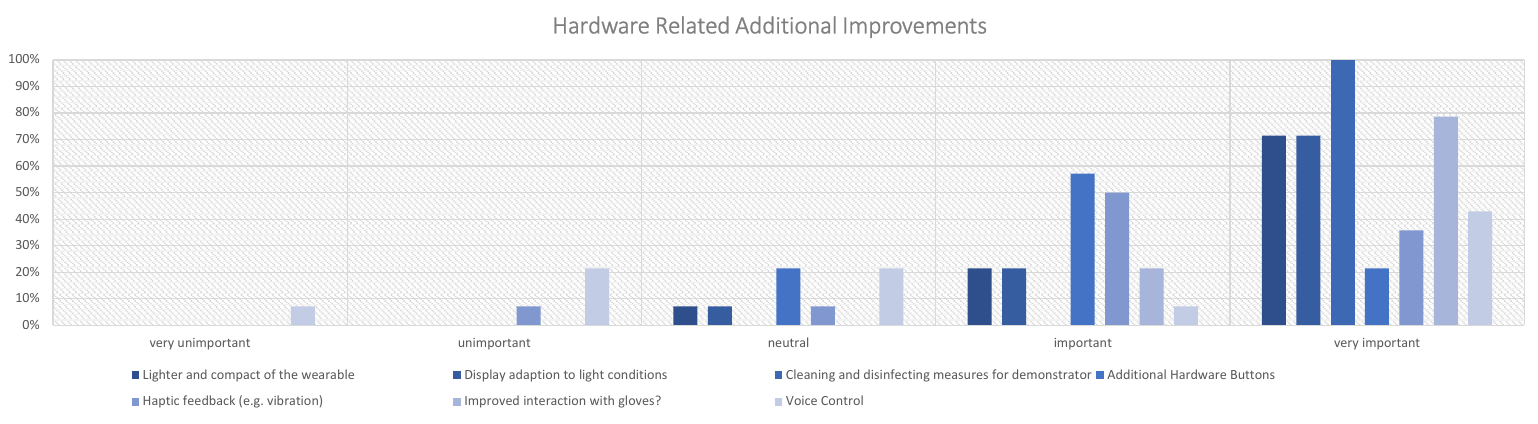}}
    \caption{\textbf{Hardware related Additional Improvements} This figure presents the, in the KDT presented, hardware related questions, which indicate a high important factor in hygience and cleaning of devices. Beside a small formfactor, a easy to clean and desinfect device is much wished from the rescue operators.}
    \label{hwfeatures}
\end{figure*}

Neutral observers watched the events to gain reactions, visual stimuli, and general impressions of the participants. The goal was to find reactions of the participants and match it with their results of the qualitative interview. An expert interview was then conducted separately with the respective testers to gain their impressions of the test. Here both participants were separated into two interview rooms, to allow personal interviews. After the interview, the hardware, embedded in this project, and the situation recognition were extensively evaluated. The end devices were provided in 3D-printed housings so that they could be actively evaluated on the body. The goal here was to provide the rescue operators the feeling of wearing the wrist-worn devices during rescue operations. Finally, a survey with twenty-three questions was provided, combining multiple choice, free text, input and Likert-scaled response options on the hardware and 3D printing. The group of participants included emergency paramedics and rescue operators aged eighteen and over. The call to test was made using the central software of the Siegen rescue station and communication with the DRK-Siegen using an invitation letter and poster.

\begin{figure}
    \centerline{\includegraphics[width=250px]{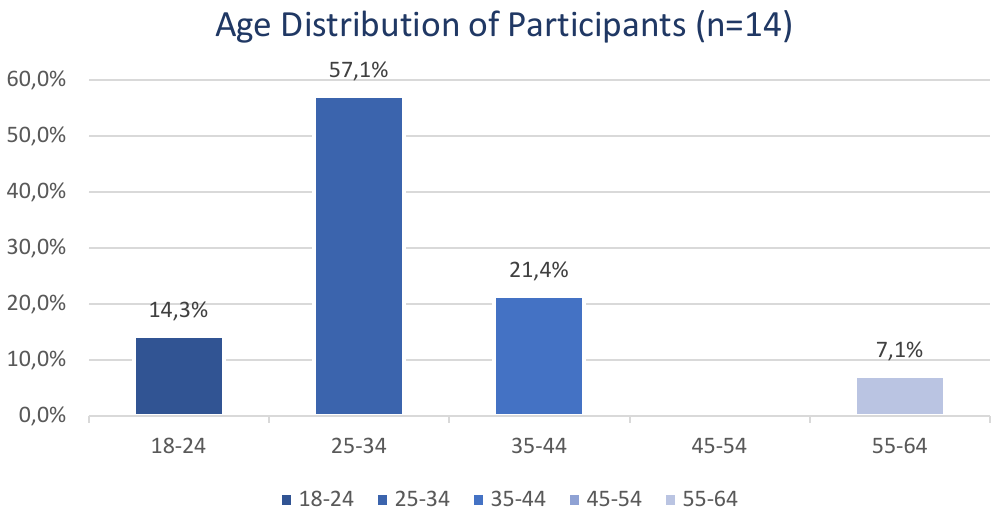}}
    \caption{\textbf{Age distribution} of the 14 participants of the KDT Testing.}
    \label{agedistribution}
\end{figure}

\subsection{Statistical Analysis}

The, in this paper presented, results are conducted through an online-survey tool (LimeSurvey) and was subsequently analyzed using MAXQDA and Microsoft Excel. Furthermore, this survey presents a vast overview of the situation at hand, knowing that multiple rescue operations-stations might be needed to be extensivly considered to provide much more in depth-analysis. The maximum values of 100\% can be achieved, if all participants (n=14) have the same opinion on a topic. Non existing values (e.g. 0\%) are not visualized in the graphical representation, meaning, that no participants had this opinion.  

\section{Results}
In the following section, the quantitative measured results are presented in four main subsections: (a) need of digitalization, (b) integration of hardware features, (c)  integration of various features and (d) weight and wearing comfort of 3d printed cases: 

\subsection{Participants}
A total of 14 participants from two different rescue stations (German Red Cross and Firestation Siegen) took part in the KIRETT-Demonstrator Testing. From those 10 participants (71\%) were from the Fire station in Siegen and 4 test users (29\%) from the German Red Cross Department from Siegen. The level of profession within the participants extended from participants in training (29\%), over 0-2 years of work experience (14\%), to 3-5 years of work experience (14\%). Experienced participants with a work experience of 6-10 (14\%) and with more than 11 years of work experience (29\%) were also included.  The age distribution covers from 18 years to above and is presented in figure \ref{agedistribution}.

\begin{figure*} 
    \centerline{\includegraphics[width=\linewidth]{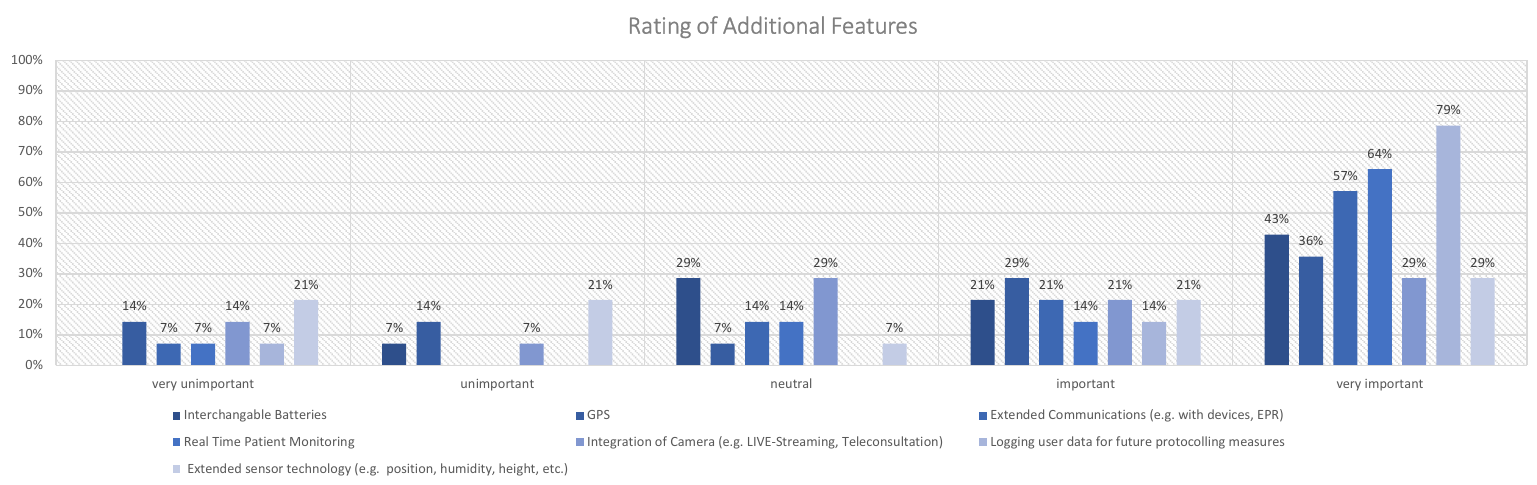}}
    \caption{\textbf{Additional features} for the KIRETT-demonstrator were identified to add further features to provide a much more domain-specific device.}
    \label{additionalfeat}
\end{figure*}

\subsection{Need of digitalization and artificial intelligence in rescue station}
Initially the participants were asked to identify, how important they see the task of digitalization in rescue operations. Most participants identified the need for digitalization in the realm of rescue operations, especially with the rising digitalization efforts in Germany (e.g. integration into the Telemedicine Act) as very relevant (relevant: 7\%, very relevant: 86\%). This clearly indicates that there is a need for modern technologies in the fast-adjusting world of healthcare. Based on this identification, follow up question about artificial intelligence (AI) has been made, covering the individuals' experience with this topic and their opinion of integrating modern technologies (e.g. AI and machine learning) into the emergency services. All of the participants indicated, that there is an importance (important: 57\%, very important: 43\%) in the integration of AI and machine-learning in their daily work, however their own personal experience with AI mostly experienced as less (little experience: 57\%) to no experience (no experience: 35\%). This shows that the need for modern technologies to aid in the treatment situation at hand is given and there should be training classes of using AI in the realm of rescue operations.

\subsection{Hardware-Related additional features}
A hardware testing was conducted during the KDT, where participants had the possibility to work actively with the devices. The goal of this was to provide a much more hands-on evaluation with the selected devices (e.g. FPGA).  The participants were asked to name possible hardware-related improvements, which could lead to better use of the devices, and additionally enhance the acceptance of the selected devices. Figure \ref{hwfeatures} described the importance level, of the participants, for the following features: 
\begin{itemize}
    \item Lighter and compact wearable design
    \item Display adaption to light conditions
    \item Cleaning and disinfecting devices
    \item Additional hardware buttons
    \item Haptic feedback (e.g. vibrations)
    \item Improved interaction with gloves
    \item Voice control
\end{itemize}

It is clearly shown, that above the 70\% mark, a lighter and compact device (71\%), with adjustable light-control options will support the user’s perception with the end devices. These remarks are assumed to be based on the given demonstrator setup. In product-development, such devices may undertake a much smaller form factor, by using microchips and microprocessors. For the, in the future used Touch-LCD, an LCD needs to be chosen where interaction with gloves (important: 79\%) needs to be made possible. Rescue Operators are constantly wearing gloves, to ensure a high hygiene standard and to secure themself and the patient of unwanted infection. Such high hygiene and disinfection level is much needed to ensure a good base for patient treatment. This is also supported by the, in figure \ref{hwfeatures} presented urge to have an easy to clean and to disinfect device. All participants (very important: 100\%) voted for this feature to be considered as very important.

\begin{figure*} 
    \centerline{\includegraphics[width=\linewidth]{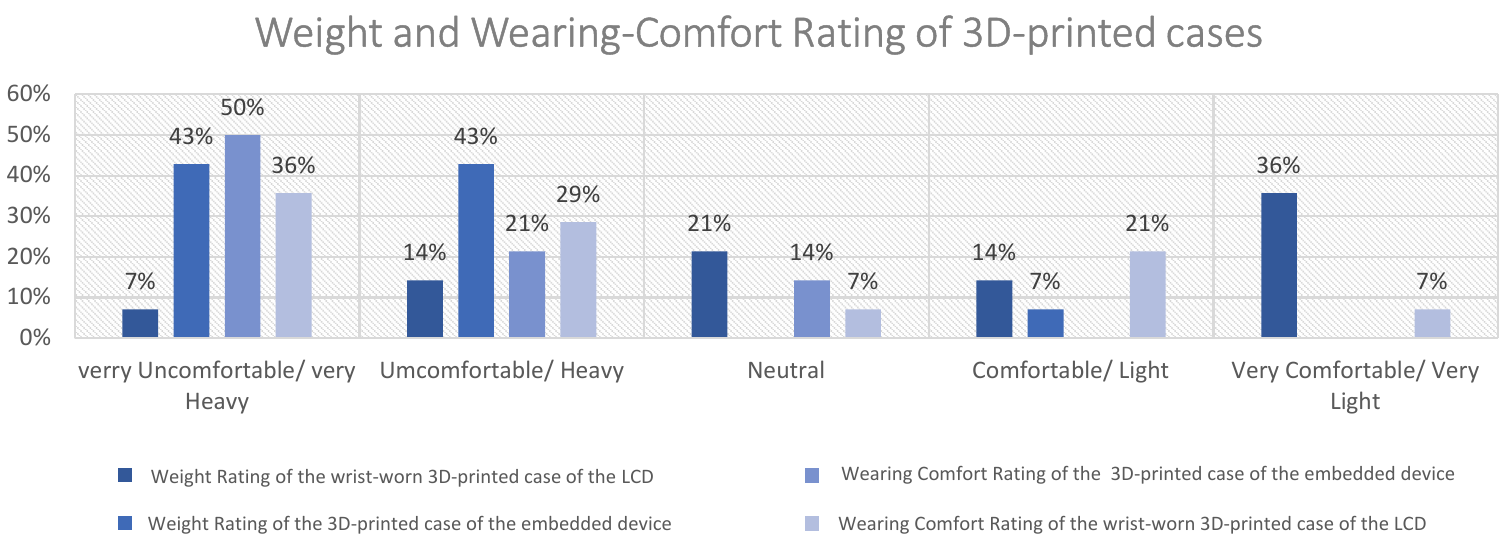}}
    \caption{\textbf{Weight and Wearing Comfort-Rating} of the 3d-printed cases.}
    \label{weight}
\end{figure*}

\subsection{Integration of various features}
Further, the participants were asked to estimate the need for various features to name possible additional developments, which would allow the domain-specific device, to achieve a higher acceptance rate within the users and to provide a more centralized device, for the rescue operators during treatment. Figure \ref{additionalfeat} describe the outcome of the survey, in which the added features were presented to the participants. The focus of this part of the survey was to identify correlations between healthcare-related digital technologies, such as real-time monitoring (Fig. \ref{additionalfeat}, extended sensor technologies for further machine-learning methods, extended communications with e.g. Electronic Health Records (EHR) or general logging of user interactions, to provide a data and action-driven protocol for the hospitals. 
The participants were asked whether extended communications with systems or national-wide electronic patient records (EPR) are categorized as important for the rescue operations. The statistics show clearly, that in rescue operations a need for communication with n-many devices is much needed (important: 21\%, very important: 57\%). Real-time monitoring was identified as the second important additional medical feature, allowing health professionals to watch the patient's vitals continuously. Many participants showed that there is the need, to track the patient's vitals (very unimportant: 7\%, neutral: 14\%, important: 15\% and very important: 64\%,) with the wearable, so fluctuations, can be directly seen and categorized for possible dangerous outcomes. 
The most needed feature is the integration of vitals and user action into the hospital. With 79\% (Fig. \ref{additionalfeat}, very important) the participants identified, that such a possibility would enhance the quality of reporting on their ends. In correlation with the real-time monitoring, reporting with time-specific vitals entries can support in retracing the decisions taken in the rescue situation. In hospital, such information can help for further treatment plans of the patient. It also can be retracted for training purposes of the rescue staff.

\subsection{Weight and Wearing Comfort of 3D-printed cases}
\begin{figure} 
    \centerline{\includegraphics[width=\linewidth]{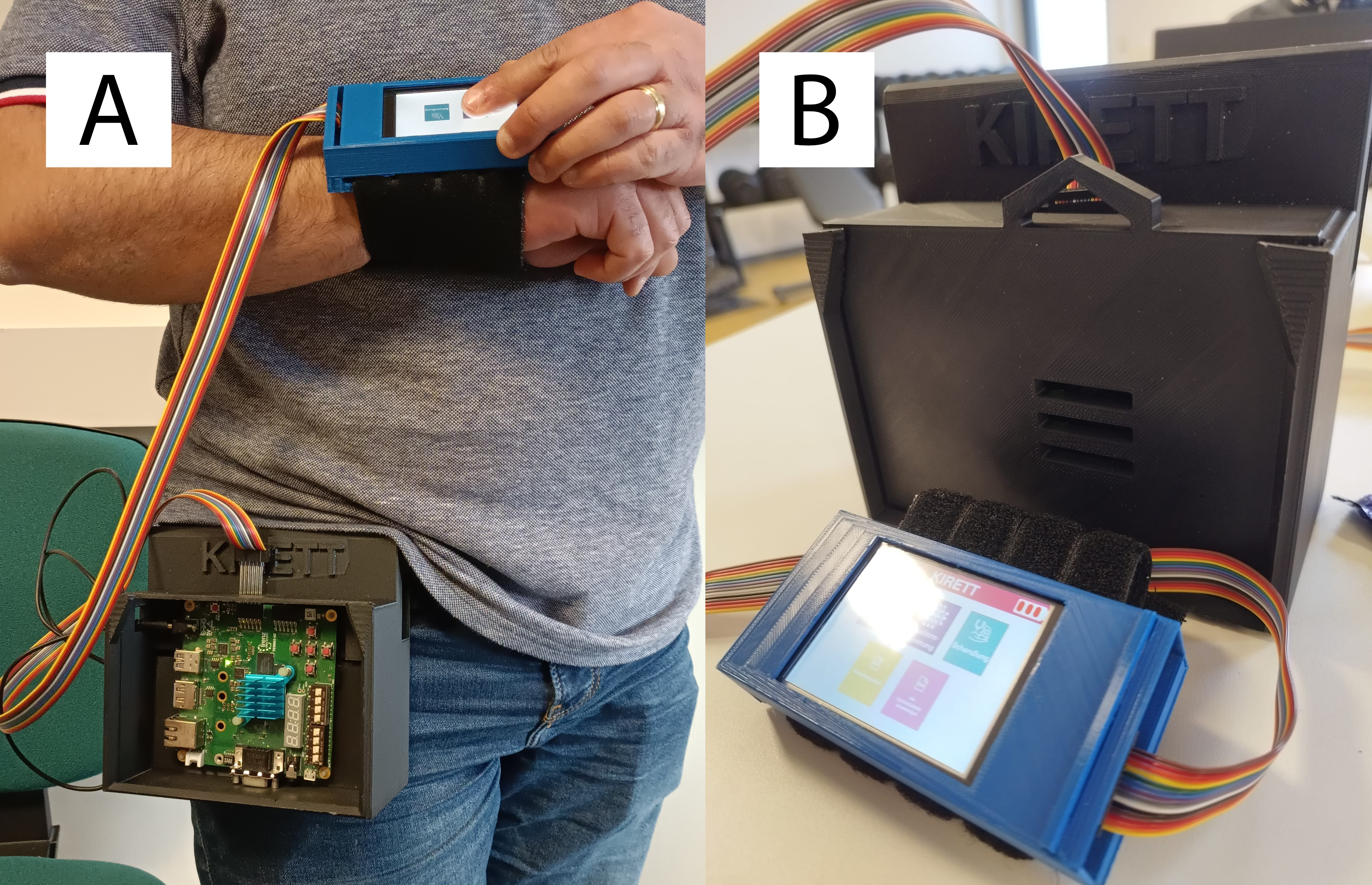}}
    \caption{\textbf{3D-Case for the KIRETT-Demonstrator} consists of 2 parts. The LCD (blue) is the wrist-worn device, which allows interaction with the software developed for the rescue operation. The black box contains, all board-relevant components, such as FPGA, Bluetooth dongle, battery pack, etc.)}
    \label{3d-case}
\end{figure}

For the KDT, 3D-printed cases for the LCD and the FPGA were designed and printed, to allow participants to use the embedded hardware actively in the testing environment. Such cases allows the user to have a clearer understanding of the sizes and proportions of the devices. They have the purpose to provide the participants an opportunity to wear the devices and get a hands-on feeling with the demonstrator devices. Figure \ref{3d-case} shows the 3D-printed devices: (A) shows the devices worn on the body to show the proportions of the cases (Fig. \ref{3d-case}, A) and (B) shows the two developed cases with the internal devices. The black box is the case for the FPGA board and its periphery, and the blue smaller case describes the wrist-worn worn LCD-case (Fig. \ref{3d-case}, B).

In today's world, 3D printing has the potential to provide fast, robust and cheap solutions for many device cases. It is highly customizable and easy to develop, which allows projects to provide a semi-feasible casing for demonstrators. In the KIRETT project, this is mainly the case. The goal was, to develop a 3D-printed case, to provide the feeling of a complete device to support the imagination of the participants for a device-prototype. Figure \ref{weight} describes the weight and wearing-comfort questions. It clearly shows that, both case: LCD case (very uncomfortable: 36\%, uncomfortable: 29\%) and the FPGA case (very uncomfortable: 50\%, uncomfortable: 29\%,) are not comfortable in wearing. This may be the case, through its bunky design. In a real prototype-development the size and design factor need to be much smaller and slimmer. Both cases were also evaluated in their weight factor. Participants' impression on the weight (very heavy: 43\%, heavy: 43\%), shows that its heavy, bulky design fused with the devices and the battery pack is not practical for rescue operations. It may hinder the treatment. However, a development in the future, where a microchips and smaller battery packs, size factor are under consideration, such device may be much more accepted by the users. The weight of the LCD-case (very comfortable: 36\%, comfortable: 14\%) was much more accepted by the participants. This may be the reason through the smaller form-factor and weight of the device and the case. This also coincides with the in figure \ref{hwfeatures} presented smaller and lighter form-factor of the wearable devices.

In conclusion, the following findings can be identified through the survey presented in this paper: 

\begin{itemize}
    \item There is a clear need of digitalization in the realm of rescue operations, covering real-time analysis through artificial intelligence or machine-learning algorithms.
    \item The future product needs to manage a smaller and slimmer weight factor, to allow a higher acceptance by the end-user. 
    \item Hygiene and gloves interactions need to be extensively researched and integrated.
    \item Real-time monitoring needs to be established through all connected devices. 
    \item The possibility to support reporting with real-time vitals and user action can be a highly demanding feature. 
    \item 3D-Case printing has its potential for fast prototyping and can be used, if the above-mentioned features and microchip-size devices can be provided. 
\end{itemize}

\section{Discussion}
This study presents the quantitative evaluation of the two-day evaluation of the KIRETT wearable in emergency medical services, with the aim of analyzing the needs of emergency medical personnel. The results provide valuable insights into the potential role of modern technology in providing fast and accurate medical treatment in emergency situations. The key finding of this study is the clear demand and need for digitalization in emergency services, with 85\% of evaluation participants emphasizing this need. This underlines the growing importance of digital solutions to ensure effective and efficient patient care. In particular, the possibility of real-time analysis and treatment recommendations is seen as crucial.

Another important aspect highlighted in the study is the need for high standards of hygiene when using wearables in the emergency services. All participants emphasized the importance of adhering to such standards, indicating the sensitivity of the healthcare sector to infection control. The issue of ease of use also emerged as crucial, particularly compatibility with medical gloves. Here, 79\% of participants emphasized the importance of ease of use under these conditions. This makes it clear that technological solutions in the emergency services must not only be functional, but also practical to use. Disinfection of the device also plays an extremely important role when working with gloves. Interestingly, real-time monitoring of vital signs (64\%) and the preparation of structured reports (79\%) were also identified as important tasks. This shows the potential of wearables not only to collect data, but also to process it in a form that is relevant for treatment. Data transmission and transfer in the hospital could be groundbreaking here.

Overall, the results of the study underline the urgent need for accelerated digitalization in the healthcare sector, especially in the field of emergency medicine. By integrating information and communication technologies, emergency services can better respond to the needs of patients and ensure seamless care. KIRETT's research findings provide a valuable contribution to the use of digital technology in emergency services and highlight the enormous potential to improve the quality and efficiency of medical care in emergency situations. However, it remains important to keep an eye on both technological developments and the needs of users to ensure that wearables actually offer added value in real-life use.

\section{Limitations}
The presented survey of the KIRETT-Demonstrator-Testing (KDT) has some limitations. The participants (n=14) of the study are mainly focused on the area around Siegen and represent only two rescue stations in the area. Interesting would be, if such testing would result in similar outcomes when the participants' amount will be increased. Based on that, does the results here has a strong representation of the needs of the rescue stations in Siegen, Germany. International tendencies may differ.

\section{Conclusion}
This paper aimed, to analyze the quantitative evaluation of the KIRETT-Demonstrator-Testing (KDT) and showed, that rescue operators require modern technologies, like artificial intelligence and machine-learning algorithms to support their daily tasks. In today's fast-paced healthcare environment, there's a critical demand for expedited patient-centric services. This underscores the necessity for enhanced digitalization within emergency medical services. This entails real-time patient health monitoring and structured reporting to hospitals, facilitating seamless post-treatment care. Embracing informatics and digitalization is imperative for swift and improved healthcare delivery.\\

\textbf{Material and methods:} This study was conducted as a combined (quantitative and qualitative) testing, in the city of Siegen in Germany, covering rescue operators and emergency operators in Siegen in Germany. The testing was made through live hands-on testing with the devices with a follow-up qualitative interview and a quantitative survey. In this paper, the focus is on the quantitative results of this testing. \\

\textbf{Results:} In total 14 participants took part in the live-testing of the KIRETT-demonstrator, covering 90 minutes of hands-on work with the devices, and follow-up interviews and surveys. This paper shows, that there is a clear need of digitalization (85\%) in the field of rescue services covering real-time analysis and treatment recommendations. Maintaining high hygiene standards (100\%)  with the devices and easy usability with medical gloves (79\%) are highly important features, which need to be taken into consideration. Real-time vitals monitoring (64\%) and the preparation of structured reporting (79\%) were documented as interesting tasks. \\

\textbf{Conclusion:} 
 The need of real-time monitoring and reporting measures to post-treatment stations (e.g. hospital) shows, that treatment recommendation and situation detection, are not only needed for sufficient treatment, but also previous reports have an important impact of the patient's treatment. Its observing measures and post-support in hospitals have a huge factor on the patient health outcome. In the future, the treatment recommendation in detail needs to be evaluated.

\bibliographystyle{unsrt}  
\bibliography{templateArXiv}

\end{document}